\documentclass[10pt,twocolumn,letterpaper]{article}

\usepackage{cvpr}
\usepackage{times}
\usepackage{epsfig}
\usepackage{graphicx}
\usepackage{amsmath}
\usepackage{amssymb}

\usepackage{multicol}
\usepackage{balance}
\usepackage{tabto,enumitem}

\makeatletter
\renewcommand{\paragraph}{%
  \@startsection{paragraph}{4}%
  {\z@}{1ex \@plus 1ex \@minus .2ex}{-1em}%
  {\normalfont\normalsize\bfseries}%
}
\makeatother

\makeatletter
\newcommand{\printfnsymbol}[1]{%
  \textsuperscript{\@fnsymbol{#1}}%
}
\makeatother

\newcommand\myindent{1.5in}
\newcommand\indentitem{\tabto{\myindent}$\bullet$\hspace{\labelsep}}

\renewcommand{\ie}{\textit{i.e.}\@\xspace}
\renewcommand{\eg}{\textit{e.g.}\@\xspace}

\usepackage[breaklinks=true,bookmarks=false]{hyperref}

\cvprfinalcopy 

\def\cvprPaperID{****} 

\begin{document}

\title{Graph-guided Architecture Search for Real-time Semantic Segmentation}

\author{
Peiwen Lin$^{1, }$\thanks{The first two authors contributed equally to this paper.}~~~~~
Peng Sun$^{2, }$\printfnsymbol{1}~~~~~
Guangliang Cheng$^{1}$~~~~~
Sirui Xie$^{3}$~~~~~
Xi Li$^{2,}$\thanks{Corresponding Author}~~~~~
Jianping Shi$^{1, }$\printfnsymbol{2} \\
\\
\and
$^{1}$SenseTime Research~~~~$^{2}$Zhejiang University~~~~$^{3}$University of California, Los Angeles~~~~
\and
{\tt\small 
\{linpeiwen,chengguangliang,shijianping\}@sensetime.com}\\
{\tt\small 
\{sunpeng1996,xilizju\}@zju.edu.cn} ~~~
{\tt\small srxie@ucla.edu}
}

\maketitle

\begin{abstract}
   Designing a lightweight semantic segmentation network often requires researchers to find a trade-off between performance and speed, which is always empirical due to the limited interpretability of neural networks. In order to release researchers from these tedious mechanical trials, we propose a Graph-guided Architecture Search (GAS) pipeline to automatically search real-time semantic segmentation networks. Unlike previous works that use a simplified search space and stack a repeatable cell to form a network, we introduce a novel search mechanism with a new search space where a lightweight model can be effectively explored through the cell-level diversity and latency-oriented constraint. Specifically, to produce the cell-level diversity, the cell-sharing constraint is eliminated through the cell-independent manner. Then a graph convolution network (GCN) is seamlessly integrated as a communication mechanism between cells. Finally, a latency-oriented constraint is endowed into the search process to balance the speed and performance. Extensive experiments on Cityscapes and CamVid datasets demonstrate that GAS achieves the new state-of-the-art trade-off between accuracy and speed. In particular, on Cityscapes dataset, GAS achieves the new best performance of 73.5\% mIoU with speed of 108.4 FPS on Titan Xp.
\end{abstract}

\begin{figure}[t]
\centering
\includegraphics[width=8.2cm]{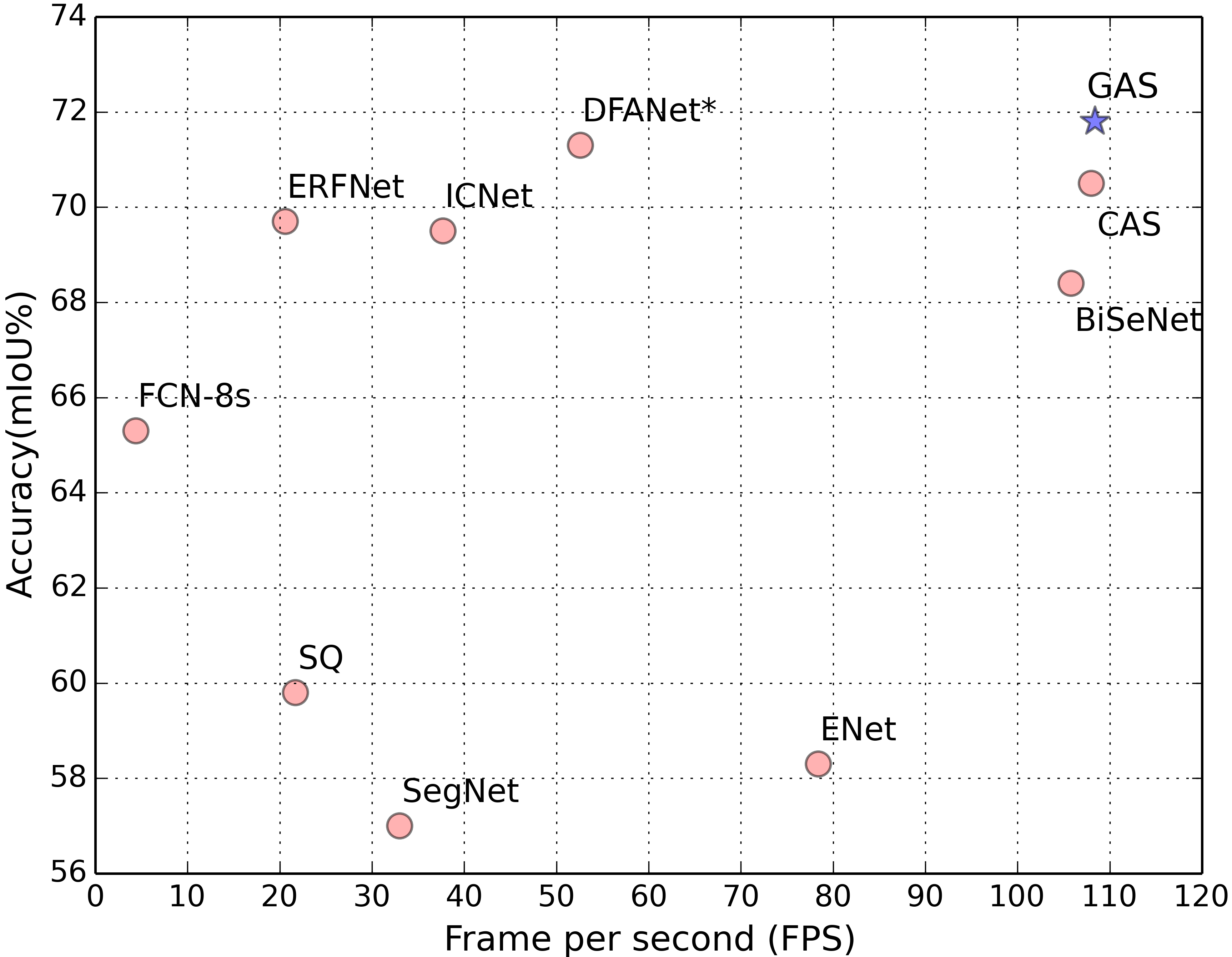}
\caption{The inference speed and mIoU for different networks on the Cityscapes test set with only fine training data. Our GAS achieves the state-of-the-art trade-off between speed and performance. The Mark $*$ denotes the speed is remeasured on Titan Xp. \textbf{Best viewed in color.}}

\label{fig:plot}
\end{figure}

\section{Introduction}

\par As a fundamental topic in computer vision, semantic segmentation \cite{long2015fully,zhao2017pyramid,chen2018deeplab,DBLP:journals/corr/ChenPSA17} aims at predicting pixel-level labels for images. Leveraging the strong ability of CNNs \cite{corr_SimonyanZ14a,conf_cvpr_HeZRS16,journals_corr_HuangLW16a, DBLP:conf/cvpr/Chollet17}, many works have achieved remarkable performance on public semantic segmentation benchmarks \cite{cordts2016cityscapes,everingham2015pascal,Camvid}. To pursue higher accuracy, state-of-the-art models become increasingly larger and deeper, and thus require high computational resources and large memory overhead, which makes it difficult to deploy on resource-constrained platforms, such as mobile devices, robotics, self-driving cars, etc.

Recently, many researches have focused on designing and improving CNN models with light computation cost and satisfactory segmentation accuracy. For example, some works \cite{badrinarayanan2017segnet,paszke2016enet} reduce the computation cost via the pruning algorithms, and ICNet \cite{zhao2018icnet} uses an image cascade network to incorporate multi-resolution inputs. BiSeNet \cite{DBLP:conf/eccv/YuWPGYS18} and DFANet \cite{li2019dfanet} utilize a light-weight backbone to speed up, and is equipped with a well-designed feature fusion or aggregation module to remedy the accuracy drop. To achieve such design, researchers acquire expertise in architecture design through enormous trial and error to carefully balance the accuracy and resource-efficiency.

To design more effective segmentation networks, some researchers have explored automatically neural architecture search (NAS) methods \cite{DBLP:conf/iclr/LiuSY19,zoph2018learning,negrinho2017deeparchitect,krause2017dynamic,DBLP:conf/icml/PhamGZLD18,cai2018proxylessnas,DBLP:conf/iclr/XieZLL19} and achieved excellent results. For example, Auto-Deeplab \cite{DBLP:journals/corr/abs-1901-02985} searches cell structures and the downsampling strategy together in the same round. CAS \cite{zhang2019customizable} searches an architecture with customized resource constraint and a multi-scale module which has been widely used in semantic segmentation field \cite{chen2018deeplab,zhao2017pyramid}.

\begin{figure}[t]
\centering
\includegraphics[width=7cm]{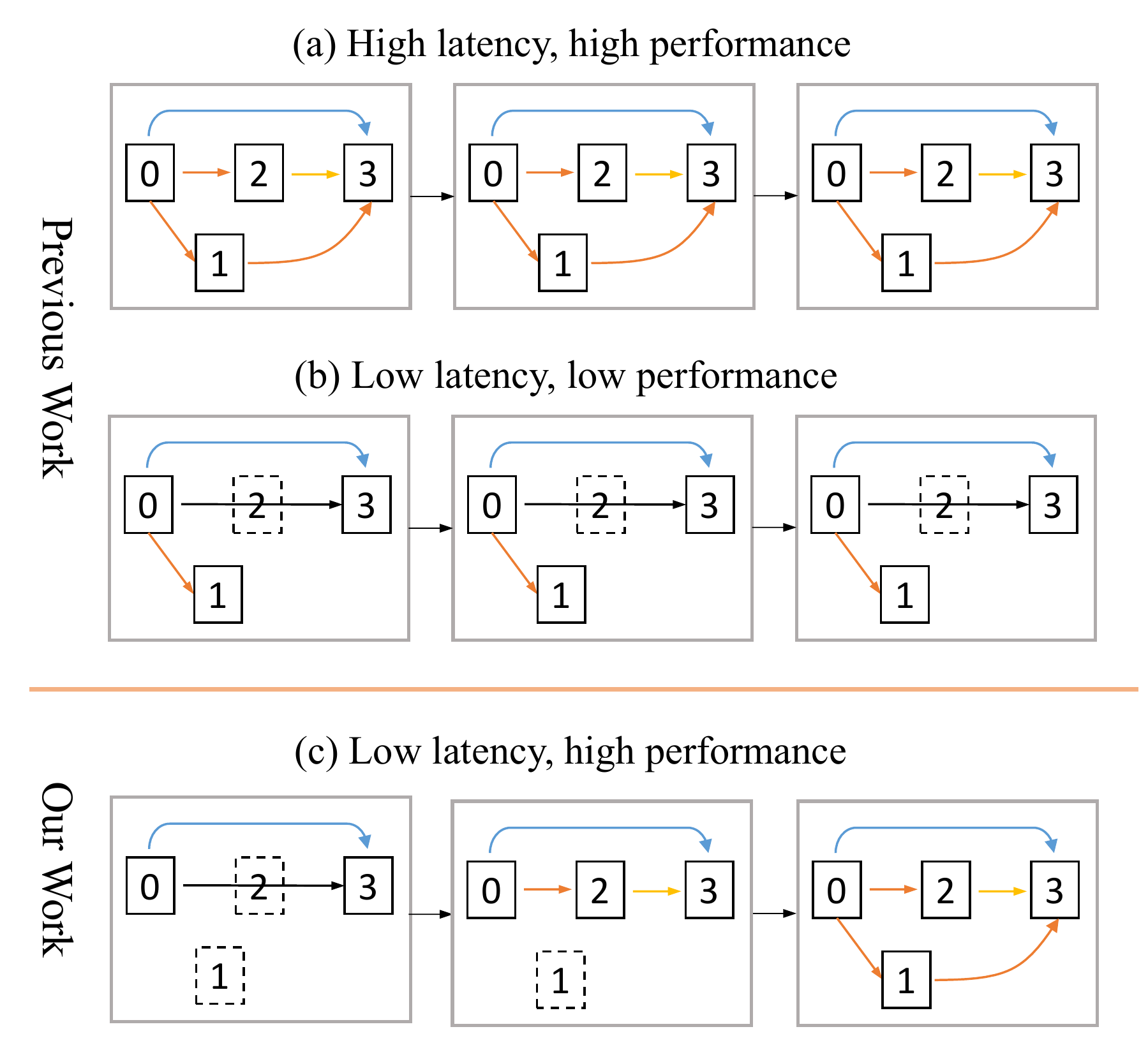}
\caption{(a) The network stacked by complicated cells results in high latency and high performance. (b) The network stacked by simple cells leads to low latency and low performance. (c) The cell diversity strategy, \ie~each cell possesses own independent structure, can flexibly construct the high accuracy lightweight network. \textbf{Best viewed in color.}}

\label{fig:flowchart}
\end{figure}

Particularly, CAS has achieved state-of-the-art segmentation performance in mobile setting~\cite{zhao2018icnet,li2019dfanet,DBLP:conf/eccv/YuWPGYS18}. Like the general NAS methods, such as ENAS \cite{DBLP:conf/icml/PhamGZLD18}, DARTS \cite{DBLP:conf/iclr/LiuSY19} and SNAS \cite{DBLP:conf/iclr/XieZLL19}, CAS also searches for two types of cells (\ie~normal cell and reduction cell) and then repeatedly stacks the identical cells to form the whole network. This simplifies the search process, but also increases the difficulties to find a good trade-off between performance and speed due to the limited cell diversity. As shown in Figure \ref{fig:flowchart}(a), the cell is prone to learn a complicated structure to pursue high performance without any resource constraint, and the whole network will result in high latency. When a low-computation constraint is applied, the cell structures tend to be over-simplified as shown in Figure \ref{fig:flowchart}(b), which may not achieve satisfactory performance.

Different from the traditional search algorithms with simplified search space, in this paper, we propose a novel search mechanism with new search space, where a lightweight model with high performance can be fully explored through the well-designed cell-level diversity and latency-oriented constraint. On one hand, to encourage the cell-level diversity, we make each cell structure independent, and thus the cells with different computation cost can be flexibly stacked to form a lightweight network in Figure \ref{fig:flowchart}(c). In this way, simple cells can be applied to the stage with high computation cost to achieve low latency, while complicated cells can be chosen in deep layers with low computation for high accuracy. On the other hand, we apply a real-world latency-oriented constraint into the search process, through which the searched model can achieve better trade-off between the performance and latency.

However, simply endowing cells with independence in exploring its own structures enlarges the search space and makes the optimization more difficult, which causes accuracy degradation as shown in Figure \ref{fig:effi_gcn}(a) and Figure \ref{fig:effi_gcn}(b). To address this issue, we incorporate a Graph Convolution Network (GCN) \cite{kipf2016semi} as the communication deliverer between cells. Our idea is inspired by \cite{minsky1988society} that different cells can be treated as multiple agencies, whose achievement of social welfare may require communication between them. Specifically, in the forward process, starting from the first cell, the information of each cell is propagated to the next adjacent cell with a GCN. Our ablation study exhibits that this communication mechanism tends to guide cells to select less-parametric operations, while achieving the satisfactory accuracy. We name the method as Graph-guided Architecture Search (GAS).

We conduct extensive experiments on the standard Cityscapes \cite{cordts2016cityscapes} and CamVid \cite{Camvid} benchmarks. Compared to other real-time methods, our method locates in the top-right area in Figure \ref{fig:plot}, which is the state-of-the-art trade-off between the performance and latency.

The main contributions can be summarized as follows:
\begin{itemize}
    		\item We propose a novel search framework, for real-time semantic segmentation, with a new search space in which a lightweight model with high performance can be effectively explored.
    		
    		\item We integrate the graph convolution network seamlessly into neural architecture search as a communication mechanism between independent cells.
  
       	    \item The lightweight segmentation network searched with GAS is customizable in real applications. Notably, GAS has achieved 73.5\% mIoU on the Cityscapes test set and 108.4 FPS on NVIDIA Titan Xp with one $769 \times 1537$ image.

\end{itemize}

\begin{figure*}
\begin{center}
\includegraphics[width=6.8in]{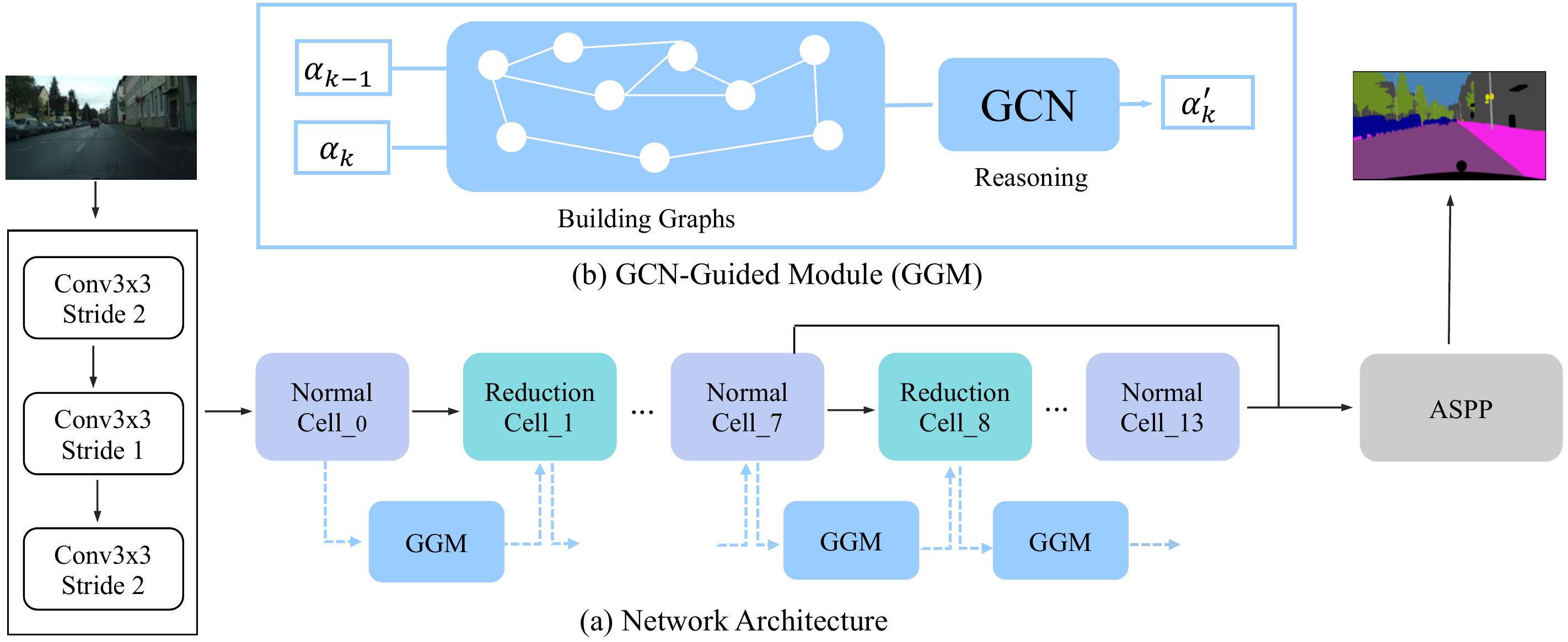}
\end{center}
\caption{Illustration of our Graph-Guided Network Architecture Search. In reduction cells, all the operations adjacent to the input nodes are of stride two. (a) The backbone network, it's stacked by a series of independent cells. (b) The GCN-Guided Module (GGM), it propagates information between adjacent cells. $\alpha_{k}$ and $\alpha_{k-1}$ represent the architecture parameters for cell $k$ and cell $k-1$, respectively, and $\alpha_{k}^{'}$ is the updated architecture parameters by GGM for cell $k$. The dotted lines indicate GGM is only utilized in the search progress. \textbf{Best viewed in color.}}
\label{fig:ggn_network}
\end{figure*}

\section{Related Work}

\textbf{Semantic Segmentation Methods} FCN~\cite{long2015fully} is the pioneer work in semantic segmentation. To improve the segmentation performance, some remarkable works have utilized various heavy backbones \cite{corr_SimonyanZ14a,conf_cvpr_HeZRS16,journals_corr_HuangLW16a,DBLP:conf/cvpr/Chollet17} or effective modules to capture multi-scale context information~\cite{zhao2017pyramid,DBLP:journals/corr/ChenPSA17,DBLP:conf/eccv/ChenZPSA18}. These outstanding works are designed for high-quality segmentation, which is inapplicable to real-time applications. In terms of efficient segmentation methods, there are two mainstreams. One is to employ relatively lighter backbone (\eg~ENet~\cite{paszke2016enet}) or introduce some efficient operations (depth-wise dilated convolution). DFANet~\cite{li2019dfanet} utilizes a lightweight backbone to speed up and equips with a cross-level feature aggregation module to remedy the accuracy drop. Another is based on multi-branch algorithm that consists of more than one path. For example, ICNet~\cite{zhao2018icnet} proposes to use the multi-scale image cascade to speed up the inference. BiSeNet~\cite{DBLP:conf/eccv/YuWPGYS18} decouples the extraction for spatial and context information using two paths.

\par \textbf{Neural Architecture Search} Neural Architecture Search (NAS) aims at automatically searching network architectures. Most existing architecture search works are based on either reinforcement learning~\cite{DBLP:conf/iclr/ZophL17,DBLP:journals/corr/abs-1812-05285} or evolutionary algorithm~\cite{DBLP:journals/corr/abs-1802-01548,DBLP:journals/corr/abs-1808-00193}. Though they can achieve satisfactory performance, they need thousands of GPU hours. To solve this time-consuming problem, one-shot methods \cite{bender2018understanding,brock2017smash} have been developed to greatly solve the time-consuming problem by training a parent network from which each sub-network can inherit the weights. They can be roughly divided into cell-based and layer-based methods according to the type of search space. For cell-based methods, ENAS~\cite{DBLP:conf/icml/PhamGZLD18} proposes a parameter sharing strategy among sub-networks, and DARTS~\cite{DBLP:conf/iclr/LiuSY19} relaxes the discrete architecture distribution as continuous deterministic weights, such that they could be optimized with gradient descent. SNAS~\cite{DBLP:conf/iclr/XieZLL19} proposes novel search gradients that train neural operation parameters and architecture distribution parameters in the same round of back-propagation. What's more, there are also some excellent works \cite{chen2019progressive,noy2019asap} to reduce the difficulty of optimization by decreasing gradually the size of search space.~For layer-based methods, FBNet~\cite{wu2019fbnet}, MnasNet~\cite{tan2019mnasnet}, ProxylessNAS~\cite{cai2018proxylessnas} use a multi-objective search approach that optimizes both accuracy and real-world latency. 

~In the field of semantic segmentation, DPC~\cite{DBLP:conf/nips/ChenCZPZSAS18} is the pioneer work by introducing meta-learning techniques into the network search problem. Auto-Deeplab \cite{DBLP:journals/corr/abs-1901-02985} searches cell structures and the downsampling strategy together in the same round. More recently, CAS \cite{zhang2019customizable} searches an architecture with customized resource constraint and a multi-scale module which has been widely used in semantic segmentation field. And \cite{nekrasov2019fast} over-parameterises the architecture during the training via a set of auxiliary cells using reinforcement learning. Recently, NAS also has been used in object detection, such as NAS-FPN \cite{nasfpn}, DetNAS \cite{chen2019detnas} and Auto-FPN \cite{Xu_2019_ICCV_Auto-FPN}.

\par \textbf{Graph Convolution Network}~Convolutional neural networks on graph-structure data is an emerging topic in deep learning research.~Kipf \cite{kipf2016semi} presents a scalable approach for graph-structured data that is based on an efficient variant of convolutional neural networks which operate directly on graphs, for better information propagation.~After that, Graph Convolution Networks (GCNs) \cite{kipf2016semi} is widely used in many domains, such as video classification \cite{wang2018videos} and action recognition \cite{stgcn2018aaai}. In this paper, we apply the GCNs to model the relationship of adjacent cells in network architecture search.

\section{Methods}
As shown in Figure~\ref{fig:ggn_network}, GAS searches for, with GCN-Guided module (GGM), an optimal network constructed by a series of independent cells. In the search process, we take the latency into consideration to obtain a network with computational efficiency. This search problem can be formulated as:
\begin{equation}\label{equ:loss}
\mathop {\min }\limits_{a \in \mathcal{A}} L_{val} + \beta * L_{lat}
\end{equation}
where $\mathcal{A}$ denotes the search space, $L_{val}$ and $L_{lat}$ are the validation loss and the latency loss, respectively. Our goal is to search an optimal architecture ${a \in \mathcal{A}}$ that achieves the best trade-off between the performance and speed.

In this section, we will describe three main components in GAS: 1) Network Architecture Search; 2) GCN-Guided Module; 3) Latency-Oriented Optimization.

\subsection{Network Architecture Search}

\par As shown in Figure \ref{fig:ggn_network}(a), the whole backbone takes an image as input which is first filtered with three convolutional layers followed by a series of independent cells. The ASPP~\cite{chen2018deeplab} module is subsequently used to extract the multi-scale context for the final prediction.

\par A cell is a directed acyclic graph (DAG) as shown in Figure \ref{fig:cell}. Each cell has two input nodes $i_{1}$ and $i_{2}$, ${N}$ ordered intermediate nodes, denoted by $\mathcal{N} = \{x_1,...,x_N\}$, and an output node which outputs the concatenation of all intermediate nodes $\mathcal{N}$. Each node represents the latent representation (\eg~feature map) in the network, and each directed edge in this DAG represents an candidate operation (\eg~conv, pooling). 

The number of intermediate nodes ${N}$ is 2 in our work. Each intermediate node takes all its previous nodes as input. In this way, ${x_1}$ has two inputs $I_{1} = \{i_1, i_2\}$ and node ${x_2}$ takes $I_{2} = \{i_1, i_2, x_1\}$ as inputs. The intermediate nodes $x_{i}$ can be calculated by:

\begin{figure}[t]
\centering
\includegraphics[width=1.2in]{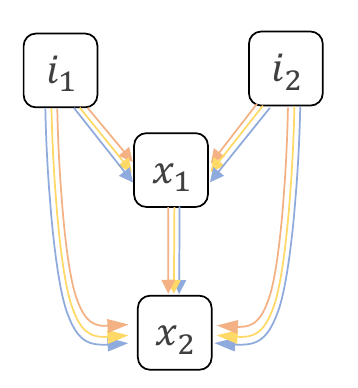}
\caption{ The structure of cell in our GAS. Each colored edge represents one candidate operation.}
\label{fig:cell}
\end{figure}

\begin{equation}
x_{i}  \! =  \! \sum_{c \in I_{i}} \widetilde{O}_{h,i}(c)
\end{equation}
where $\widetilde{O}_{h,i}$ is the selected operation at edge ($h$, $i$). 

To search the selected operation $\widetilde{O}_{h,i}$, the search space is represented with a set of one-hot random variables from a fully factorizable joint distribution $p(Z)$ \cite{DBLP:conf/iclr/XieZLL19}. Concretely, each edge is associated with a one-hot random variable which is multiplied as a mask to the all possible operations ${O}_{h,i}$ = ($o^{1}_{h,i}$, $o^{2}_{h,i}$, ..., $o^{M}_{h,i}$) in this edge. We denote the one-hot random variable as $Z_{h,i}$ = ($z^{1}_{h,i}$, $z^{2}_{h,i}$, ..., $z^{M}_{h,i}$) where ${M}$ is the number of candidate operations. The intermediate nodes during search process in such way are:

\begin{small}
\begin{equation}
x_{i}  \! =  \! \sum_{c \in I_{i}} \widetilde{O}_{h,i}(c) = \sum_{c \in I_{i}} \sum_{m=1}^{M} z_{h,i}^m o_{h,i}^m(c)
\end{equation}
\end{small}

To make $P(Z)$ differentiable, reparameterization \cite{maddison2016concrete} is used to relax the discrete architecture distribution to be continuous:
\begin{small}
\begin{equation}\label{equ:softmax}
Z_{h,i} \! = f_{\alpha_{h,i}}(G_{h,i}) = softmax((log \alpha_{h,i} + G_{h,i}) / \lambda)
\end{equation}
\end{small}
where $\alpha_{h,i}$ is the architecture parameters at the edge $(h,i)$, and $G_{h,i}$ = $-log(-log(U_{h,i}))$ is a vector of Gumbel random variables, $U_{h,i}$ is a uniform random variable and $\lambda$ is the temperature of softmax.

For the set of candidate operations $O$, we only use the following 8 kinds of operations to better balance the speed and performance:

\vspace{-0.06in}
\begin{itemize}[itemsep=0pt, parsep=0pt]
\footnotesize
\item 3 $\times$ 3 max pooling
\indentitem skip connection
\item 3 $\times$ 3 conv
\indentitem zero operation
\item 3 $\times$ 3 separable conv
\item 3 $\times$ 3 dilated separable conv (dilation=2)
\item 3 $\times$ 3 dilated separable conv (dilation=4)
\item 3 $\times$ 3 dilated separable conv (dilation=8)
\end{itemize}
\label{Section 3.3}
\vspace{-0.2in}

\subsection{GCN-Guided Module}

\par With cell independent to each other, the inter-cell relationship becomes very important for searching efficiently. We propose a novel GCN-Guided Module (GGM) to naturally bridge the operation information between adjacent cells. The total network architecture of our GGM is shown in Figure \ref{fig:ggn_network}(b). Inspired by \cite{wang2018videos}, the GGM represents the communication between adjacent cells as a graph and perform reasoning on the graph for information delivery. Specifically, we utilize the similarity relations of edges in adjacent cells to construct the graph where each node represents one edge in cells. In this way, the state changes for previous cell can be delivered to current cell by reasoning on this graph.

\par As stated in Section $3.1$, let $\alpha_{k}$ represents the architecture parameter matrix for the cell $k$, and the dimension of $\alpha_{k}$ is $p$ $\times$ $q$ where $p$ and $q$ represents the number of edges and the number of candidate operations respectively. Same for cell $k$, the architecture parameter $\alpha_{k-1}$ for cell $k-1$ also is a $p$ $\times$ $q$ matrix. To fuse the architecture parameter information of previous cell $k-1$ into the current cell $k$ and generate the updated $\alpha'_{k}$, we model the information propagation between cell ${k-1}$ and cell $k$ as follows:

\begin{small}
\begin{equation}\label{equ:residual1}
\alpha'_{k} =  \alpha_{k}  +  \gamma \Phi_{2}(G(\Phi_{1}(\alpha_{k-1}), Adj)) \\
\end{equation}
\end{small}
where $Adj$ represents the adjacency matrix of the reasoning graph between cells $k$ and $k-1$, and the function $G$ denotes the Graph Convolution Networks (GCNs)~\cite{kipf2016semi} to perform reasoning on the graph. $\Phi_{1}$ and $\Phi_{2}$ are two different transformations by 1D convolution. Specifically, $\Phi_{1}$ maps the original architecture parameter to embedding space and $\Phi_{2}$ transfers it back into the source space after the GCN reasoning. $\gamma$ controls the fusion of two kinds of architecture parameter information.

For the function $G$, we construct the reasoning graph between cell ${k-1}$ and cell $k$ by their similarity. Given a edge in cell $k$, we calculate the similarity between this edge and all other edges in cell $k-1$ and a softmax function is used for normalization. Therefore, the adjacency matrix $Adj$ of the graph between two adjacent cells $k$ and $k-1$ can be established by:

\begin{equation}\label{equ:adject}
Adj =  Softmax( \phi_1 (\alpha_{k}) * \phi_2 ({ \alpha_{k-1}}) ^{T} )
\end{equation}
where we have two different transformations $\phi_1$ = $\alpha_{k}{w_1}$ and $\phi_2$ = $\alpha_{k-1}{w_2}$ for the architecture parameters, and parameters $w_1$ and $w_2$ are both $q \times q$ weights which can be learned via back propagation. The result $Adj$ is a $p \times p$ matrix.

Based on this adjacency matrix $Adj$, we use the GCNs to perform information propagation on the graph as shown in Equation \ref{equ:gcn}. A residual connection is added to each layer of GCNs. The GCNs allow us to compute the response of a node based on its neighbors defined by the graph relations, so performing graph convolution is equivalent to performing message propagation on the graph.

\begin{small}
\begin{equation}\label{equ:gcn}
G(\Phi_{1}(\alpha_{k-1}), Adj) = Adj \Phi_{1}(\alpha_{k-1}) W_{k-1}^{g}  + \Phi_{1}(\alpha_{k-1})
\end{equation}
\end{small}
where the $W_{k-1}^{g}$ denotes the GCNs weight with dimension $d \times d$, which can be learned via back propagation. 

The proposed well-designed GGM seamlessly integrates the graph convolution network into neural architecture search, which can bridge the operation information between adjacent cells.

\subsection{Latency-Oriented Optimization}

To obtain a real-time semantic segmentation network, we take the real-world latency into consideration during the search process, which orients the search process toward the direction to find an optimal lightweight model. Specifically, we create a GPU-latency lookup table \cite{cai2018proxylessnas,wu2019fbnet,zhang2019customizable,tan2019mnasnet} which records the inference latency of each candidate operation. During the search process, each candidate operation $m$ at edge ($h$, $i$) will be assigned a cost $lat_{h,i}^{m}$ given by the pre-built lookup table. In this way, the total latency for cell $k$ is accumulated as:

\begin{equation}\label{equ:latency}
lat_k \! = \! \sum_{h,i}\sum_{m=1}^{M} z_{h,i}^{m} lat_{h,i}^{m}
\end{equation}
where $z_{h,i}^{m}$ is the softened one-hot random variable as stated in Section 3.1. Given an architecture $a$, the total latency cost is estimated as:

\begin{equation}\label{equ:latency2}
LAT(a) \! = \! \sum_{k=0}^{K} lat_{k}
\end{equation}
where $K$ refers to the number of cells in architecture $a$. The latency for each operation $lat_{h,i}^{m}$ is a constant and thus total latency loss is differentiable with respect to the architecture parameter $\alpha_{h,i}$. The total loss function is designed as follows:
\begin{equation}\label{equ:latency2}
L(a, w) \! = CE(a, w_a) + \beta ~log(LAT(a))
\end{equation}
where $CE(a, w_a)$ denotes the cross-entropy loss of architecture $a$ with parameter $w_a$, $LAT(a)$ denotes the overall latency of architecture $a$, which is measured in micro-second, and the coefficient $\beta$ controls the balance between the accuracy and latency. The architecture parameter $\alpha$ and the weight $w$ are optimized in the same round of back-propagation.


\section{Experiments}
 In this section, we conduct extensive experiments to verify the effectiveness of our GAS. Firstly, we compare the network searched by our method with other works on two standard benchmarks. Secondly, we perform the ablation study for the GCN-Guided Module and latency optimization settings, and close with an insight about GCN-Guided Module.

\subsection{Benchmark and Evaluation Metrics}

\paragraph{Datasets} In order to verify the effectiveness and robustness of our method, we evaluate our method on the Cityscapes \cite{cordts2016cityscapes} and CamVid~\cite{Camvid} datasets. Cityscapes \cite{cordts2016cityscapes} is a public released dataset for urban scene understanding. It contains 5,000 high quality pixel-level fine annotated images (2975, 500, and 1525 for the training, validation, and test sets, respectively) with size 1024 $\times$ 2048 collected from 50 cities. The dense annotation contains 30 common classes and 19 of them are used in training and testing. CamVid~\cite{Camvid} is another public released dataset with object class semantic labels. It contains 701 images in total, in which 367 for training, 101 for validation and 233 for testing. The images have a resolution of 960 $\times$ 720 and 11 semantic categories.

 \paragraph{Evaluation Metrics} For evaluation, we use mean of class-wise intersection over union (mIoU), network forward time (Latency), and Frames Per Second (FPS) as the evaluation metrics.

\subsection{Implementation Details}

\par We conduct all experiments using Pytorch 0.4 \cite{pytorch} on a workstation, and the inference time in all experiments is reported on one Nvidia Titan Xp GPU.

The whole pipeline contains three sequential steps: search, pretraining and finetuning. It starts with the search progress on the target dataset and obtains the light-weight architecture according to the optimized $\alpha$ followed by the ImageNet~\cite{deng2009imagenet} pretraining, and this pretrained model is subsequently finetuned on the specific dataset for 200 epochs.

In search process, the architecture contains 14 cells and each cell has $N$ = 2 intermediate nodes. With the consideration of speed, the initial channel for network is 8. For the training hyper-parameters, the mini-batch size is set to 16. The architecture parameters $\alpha$ are optimized by Adam, with initial learning rate 0.001, $\beta$ = (0.5, 0.999) and weight decay 0.0001. The network parameters are optimized using SGD with momentum 0.9, weight decay 0.001, and cosine learning scheduler that decays learning rate from 0.025 to 0.001. For gumbel softmax, we set the initial temperature $\lambda$ in equation \ref{equ:softmax} as 1.0, and gradually decrease to the minimum value of 0.03. The search time cost on Cityscapes takes approximately 10 hours with 16 TitanXP GPU.

For finetuning details, we train the network with mini-batch 8 and SGD optimizer with `poly' scheduler that decay learning rate from 0.01 to zero. Following \cite{DBLP:journals/corr/WuSH16a}, the online bootstrapping strategy is applied to the finetuning process. For data augmentation, we use random flip and random resize with scale between 0.5 and 2.0. Finally, we randomly crop the image with a fixed size for training.

For the GCN-guided Module, we use one Graph Convolution Network (GCN)~\cite{kipf2016semi} between two adjacent cells, and each GCN contains one layer of graph convolutions. The kernels size of the GCN parameters $W$ in equation \ref{equ:gcn} is 64 $\times$ 64. We set the $\gamma$ as 0.5 in equation \ref{equ:residual1} in our experiments.

\begin{table}[h]
\begin{center}
\scalebox{0.9}{
\setlength{\tabcolsep}{1mm}{
\begin{tabular}{|l|c|c|c|c|}
\hline
Method        & Input Size & mIoU (\%) & Latency(ms) & FPS \\
\hline\hline
FCN-8S \cite{long2015fully} & 512x1024  & 65.3 & 227.23 & 4.4    \\
PSPNet \cite{zhao2017pyramid}  & 713x713   & 81.2 & 1288.0 & 0.78   \\
DeepLabV3$^*$ \cite{DBLP:journals/corr/ChenPSA17} & 769x769   & 81.3 & 769.23 & 1.3    \\
\hline
SegNet  \cite{badrinarayanan2017segnet}      & 640x360   & 57.0 & 30.3   & 33     \\
ENet  \cite{paszke2016enet}         & 640x360   & 58.3 & 12.7   & 78.4   \\
SQ  \cite{treml2016speedingSQ}          & 1024x2048 & 59.8 & 46.0   & 21.7   \\
ICNet  \cite{zhao2018icnet}       & 1024x2048 & 69.5 & 26.5   & 37.7   \\
SwiftNet \cite{SwiftNet} & 1024x2048 & 75.1 & 26.2   & 38.1 \\
ESPNet \cite{mehta2018espnet}  & 1024x512  & 60.3 & 8.2    & 121.7 \\
BiSeNet   \cite{DBLP:conf/eccv/YuWPGYS18}    & 768x1536  & 68.4 & 9.52   & 105.8  \\
DFANet A$\S$  \cite{li2019dfanet}   & 1024x1024 & 71.3 & 10.0  & 100.0   \\
DFANet A$^{\dag}$ \cite{li2019dfanet} \footnote[1]  & 1024x1024 & 71.3 & 19.01  & 52.6   \\
CAS \cite{zhang2019customizable} & 768x1536  & 70.5 & 9.25   & 108.0  \\
CAS$^*$ \cite{zhang2019customizable} & 768x1536  & 72.3 & 9.25   & 108.0  \\
\hline
\textbf{GAS}       & 769x1537  & 71.8 & 9.22   & 108.4  \\
\textbf{GAS$^*$}     & 769x1537  & 73.5 & 9.22   & 108.4  \\
\hline
\end{tabular}}
}
\end{center}
\caption{Comparing results on the Cityscapes test set. Methods trained using both fine and coarse data are marked with $*$. The mark $\S$ represents the speed on TitanX, and the mark ${\dag}$ represents the speed is remeasured on Titan Xp.}
\label{Cityscapes}
\end{table}

\subsection{Real-time Semantic Segmentation Results}

In this part, we compare the model searched by GAS with other existing real-time segmentation methods on semantic segmentation datasets. The inference time is measured on an Nvidia Titan Xp GPU and the speed of other methods reported on Titan Xp GPU in CAS \cite{zhang2019customizable} are used for fair comparison. Moreover, the speed is remeasured on Titan Xp if the origin paper reports it on different GPU and is not mentioned in CAS \cite{zhang2019customizable}.

\textbf{Results on Cityscapes.} We evaluate the network searched by GAS on the Cityscapes test set. The validation set is added to train the network before submitting to Cityscapes online server. Following \cite{DBLP:conf/eccv/YuWPGYS18,zhang2019customizable}, GAS takes as an input image with size 769 $\times$ 1537 that is resized from origin image size 1024 $\times$ 2048. Overall, our GAS gets the best performance among all methods with the speed of 108.4 FPS. With only fine data and without any evaluation tricks, our GAS yields 71.8\% mIoU which is the state-of-the-art trade-off between performance and speed for real-time semantic segmentation. GAS achieves 73.5\% when the coarse data is added into the training set. The full comparison results are shown in Table \ref{Cityscapes}. Compared to BiSeNet \cite{DBLP:conf/eccv/YuWPGYS18} and CAS \cite{zhang2019customizable} that have comparable speed with us, our GAS surpasses them along multiple performance points with 3.4\% and 1.3\%, respectively. Compared to other methods such as SegNet \cite{badrinarayanan2017segnet}, ENet \cite{paszke2016enet}, SQ \cite{treml2016speedingSQ} and ICNet \cite{zhao2018icnet}, our method achieves significant improvement in speed while getting performance gain over them about 14.8\%, 13.5\%, 12.0\%, 2.3\%, respectively.

\footnotetext[1]{After merging the BN layers for DFANet, there still has a speed gap between the original paper and our measurement. We suspect that it's caused by the inconsistency of implementation platform in which DFANet has the optimized depth-wise convolution (DW-Conv). GAS also have many candidate operations using DW-Conv, so the speed of our GAS is still capable of beating it if the DW-Conv is optimized correctly like DFANet.}

\textbf{Results on CamVid.} We directly transfer the network searched on Cityscapes to Camvid to verify the transferability of GAS. Table \ref{CamVid} shows the comparison results with other methods. With input size 720 $\times$ 960, GAS achieves the 72.8\% mIoU with 148.0 FPS which is also the state-of-the-art trade-off between performance and speed, which demonstrates the superior transferability of GAS.

\begin{table}[h]
\begin{center}
\scalebox{0.9}{
\setlength{\tabcolsep}{1mm}{
\begin{tabular}{|l|c|c|c|}
\hline
Method  & ~mIoU (\%)~ & Latency(ms) & ~FPS~ \\
\hline\hline
SegNet \cite{badrinarayanan2017segnet}   & 55.6 & 34.01 & 29.4   \\
ENet \cite{paszke2016enet}  & 51.3 & 16.33 & 61.2   \\ 
ICNet \cite{zhao2018icnet} & 67.1 & 28.98 & 34.5   \\
BiSeNet  \cite{DBLP:conf/eccv/YuWPGYS18} & 65.6 &  -    & -   \\
DFANet A \cite{li2019dfanet} & 64.7 & 8.33  & 120    \\
CAS \cite{zhang2019customizable} & 71.2 & 5.92  & 169    \\
\hline
GAS   & 72.8 & 6.53  & 153.1  \\
\hline
\end{tabular}}
}
\end{center}
\caption{Results on the CamVid test set with resulotion 960 $\times$ 720. "-" indicates the corresponding result is not provided by the methods.}
\label{CamVid}
\end{table}

\subsection{Ablation Study}
\par To verify the effectiveness of each component in our framework, extensive ablation studies for the GCN-Guided Module and the latency loss are performed. In addition, we also give some insights about the role of GCN-Guided Module in the search process.

\subsubsection{Effectiveness of the GCN-Guided Module}

\par
\begin{figure}[h]
\centering
\includegraphics[width=3.3in]{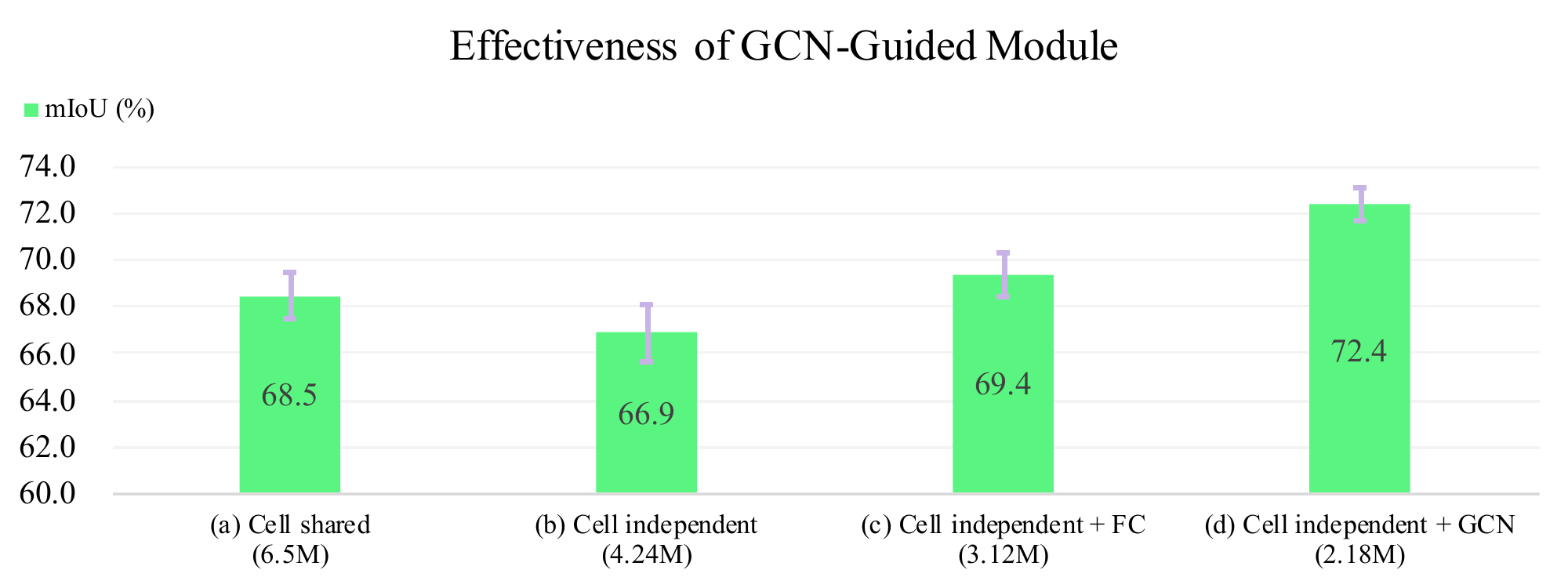}
\caption{Ablation study for the effectiveness of GCN-Guided Module on Cityscapes validation dataset. \textbf{Best viewed in color.}}
\label{fig:effi_gcn}
\end{figure}

 We propose the GCN-Guided Module (GGM) to build the connection between cells. To verify the effectiveness of the GGM, we conduct a series of experiments with different strategies: a) network stacked by shared cell; b) network stacked by independent cell; c) based on strategy-b, using fully connected layer to infer the relationship between cells; d) based on strategy-b, using GGM to infer the relationship between cells. Experimental results are shown in Figure \ref{fig:effi_gcn}. The performance reported here is the average mIoU over five repeated experiments on the Cityscapes validation set with latency loss weight $\beta$ = 0.005. The numbers below the horizontal axis are the average model size of five architectures (\eg~2.18M) and the purple line is the variance of mIoU for each strategy. Overall, with only independent cell, the performance degrades a lot due to the enlarged search space which makes optimization more difficult. This performance drop is mitigated by adding communication mechanism between cells. Especially, our GCN-Guided Module can bring about 3\% performance improvement compared to the fully-connected mechanism (\ie~setting (c)).

\paragraph{Comparison against Random Search} As discussed in \cite{DBLP:conf/uai/LiT19}, random search is a competitive baseline for hyper-parameter optimization. To further verify the effectiveness of GCN-Guided Module, we randomly sample ten architectures from the search space and evaluate them on the Cityscapes validation set with ImageNet pretrained. Specifically, we try two types of random settings in our experiments: a) fully random search without any constraint; b) randomly select the networks that meet the speed requirement about 108 FPS from the search space. The results are shown in Table \ref{randomsearch}, in which each value is the average result of ten random architectures. In summary, the network searched by GAS can achieve an excellent trade-off between performance and latency, while random search will result in high overhead without any latency constraint or low performance with latency constraint.

\begin{table}[h]
\begin{center}
\setlength{\tabcolsep}{1mm}{
\scalebox{0.9}{
\begin{tabular}{|l|c|c|}
\hline
Methods  & mIoU (\%) & FPS \\
\hline\hline
GAS  & 72.3  & 108.2      \\
Random setting (a)  & 69.6  & 61.2  \\
Random setting (b)  & 65.8  & 105.6  \\
\hline
\end{tabular}}
}
\end{center}
\caption{Comparison to random search on the Cityscapes validation set.}
\label{randomsearch}
\end{table}

\paragraph{Dimension Selection} The dimension selection of GCN weight $W$ in Equation \ref{equ:gcn} is also important, thus we conduct experiments with different GCN weight dimensions (denoted by $d$). Experimental results are shown in Table \ref{GCN_channel} in which the values are the average mIoU over five repeated experiments on the Cityscapes validation set with latency loss weight $\beta$ = 0.005. Experimental result indicates that GAS achieves the best performance when d = 64.

\begin{table}[h]
\begin{center}
\setlength{\tabcolsep}{1mm}{
\scalebox{0.9}{
\begin{tabular}{|l|c|c|}
\hline
Methods  & mIoU (\%) & FPS \\
\hline\hline
GCN with d = 16  & 71.6  & 108.6      \\
GCN with d = 32  & 71.8  & 102.2      \\
GCN with d = 64  & 72.4  & 108.4      \\
GCN with d = 128 & 72.1  & 104.1      \\
GCN with d = 256 & 71.5  & 111.2      \\
\hline
\end{tabular}}
}
\end{center}
\caption{Ablation study for different GCN weight dimensions of the GCN-Guided Module.}
\label{GCN_channel}
\end{table}

\paragraph{Reasoning Graph} For GCN-Guided Module, in addition to the way described in Section 3.2, we also try another way to construct the reasoning graph. Specifically, we treat each candidate operation in a cell as a node in the reasoning graph. Given the architecture parameter $\alpha_{k}$ for cell $k$ with dimension $p \times q$, we first flatten the $\alpha_{k}$ and $\alpha_{k-1}$ to the one dimensional vector $\alpha'_{k}$ and $\alpha'_{k-1}$, and then perform matrix multiplication to get adjacent matrix $Adj = \alpha'_{k} (\alpha'_{k-1})^{T}$. Different from the ``edge-similarity'' reasoning graph in Section 3.2, we call this graph ``operation-identity'' reasoning graph. We conduct the comparison experiment for two types of graphs on the Cityscapes validation set under the same latency loss weight $\beta$ = 0.005, the comparison results are shown in Table \ref{reasongraph}.

\begin{table}[h]
\begin{center}
\setlength{\tabcolsep}{1mm}{
\scalebox{0.9}{
\begin{tabular}{|l|c|c|c|}
\hline
Reasoning Graph  &  mIoU (\%)  & FPS \\
\hline\hline
Edge-similarity      & 72.4 & 108.4   \\
Operation-identity & 70.9 & 102.2   \\
\hline
\end{tabular}}
}
\end{center}
\caption{The comparison results for reasoning graph for edges and operations.}
\label{reasongraph}
\end{table}

Intuitively, the ``operation-identity'' way provides more fine-grained information about operation selection for other cells, while it also breaks the overall properties of an edge, and thus doesn't consider the other operation information at the same edge when making decision. After visualizing the network, we also found that the ``operation-identity'' reasoning graph tends to make cell select the same operation for all edge, which increases the difficulty of trade-off between performance and latency. This can also be verified from result in Table \ref{reasongraph}. So we choose the ``edge-similarity'' way to construct the reasoning graph as described in Section 3.2.

\paragraph{Network Visualization} We illustrate the network structure searched by GAS in the supplementary material. An interesting observation is that the operations selected by GAS with GGM have fewer parameters and less computational complexity than GAS without GGM, where more dilated or separated convolution kernels are preferred. This exhibits the emergence of concept of burden sharing in a group of cells when they know how much others are willing to contribute.

\subsubsection{Effectiveness of the Latency Constraint}

\begin{figure}[h]
\centering
\includegraphics[width=3.0in]{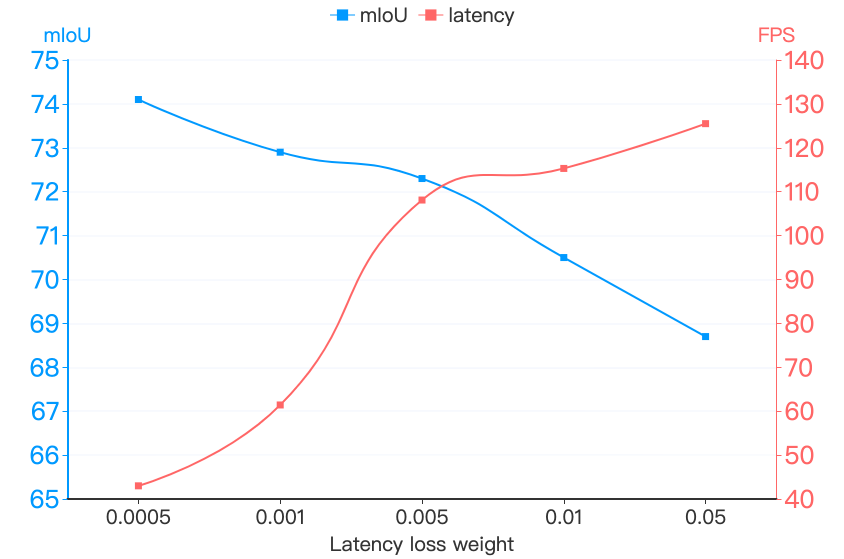}
\caption{The validation accuracy on Cityscapes dataset for different latency constraint. \textbf{Best viewed in color.}}
\label{fig:compare}
\end{figure}

As mentioned above, GAS provides the ability to flexibly achieve a superior trade-off between the performance and speed with the latency-oriented optimization. We conducted a series of experiments with different loss weight $\beta$ in Equation \ref{equ:latency2}. Figure \ref{fig:compare} shows the variation of mIoU and latency as $\beta$ changes. With smaller $\beta$, we can obtain a model with higher accuracy, and vice-versa. When the $\beta$ increases from 0.0005 to 0.005, the latency decreases rapidly and the performance is slightly falling. But when $\beta$ increases from 0.005 to 0.05, the performance drops quickly while the latency decline is fairly limited. Thus in our experiments, we set $\beta$ as 0.005. We can clearly see that the latency-oriented optimization is effective for balancing the accuracy and latency.

\subsubsection{Analysis of the GCN-Guided Module} 

One concern is about what kind of role does GCN play in the search process. We suspect that its effectiveness is derived from the following two aspects: 1) to search a light-weight network, we do not allow the cell structures to share with each other to encourage structure diversity. Apparently, learning cell independently makes the search more difficult and does not guarantee better performance, thus the GCN-Guided Module can be regraded as a regularization term to the search process. 2) We have discussed that $p(Z)$ is a fully factorizable joint distribution in above section. As shown in Equation \ref{equ:softmax}, $p(Z_{h,i})$ for current cell becomes a conditional probability if the architecture parameter $\alpha_{h,i}$ depends on the probability $\alpha_{h,i}$ for previous cell. In this case, the GCN-Guided Module plays a role to model the condition in probability distribution $p(Z)$.

\section{Conclusion \& Discussion}
\par In this paper, a novel Graph-guided architecture search (GAS) framework is proposed to tackle the real-time semantic segmentation task. Different from the existing NAS approaches that stack the same searched cell into a whole network, GAS explores to search different cell architectures and adopts the graph convolution network to bridge the information connection among cells. In addition, a latency-oriented constraint is endowed into the search process for balancing the accuracy and speed. Extensive experiments have demonstrated that GAS performs much better than the state-of-the-art real-time segmentation approaches. 
\par In the future, we will extend the GAS to the following directions: 1) we will search networks directly for the segmentation and detection tasks without retraining. 2) we will explore some deeper research on how to effectively combine the NAS and the graph convolution network.

\paragraph{Acknowledgement} This paper is carried out at SenseTime Research in Beijing, China, and is supported by key scientific technological innovation research project by Ministry of Education, Zhejiang Provincial Natural Science Foundation of China under Grant LR19F020004, Zhejiang University K.P.Chao's High Technology Development Foundation.

{\small
\bibliographystyle{ieee_fullname}
\bibliography{egbib}
}

\clearpage 
\appendix
\noindent\textbf{\Large Supplemental Material}

\section{Network Visualization}\label{visualization}
\begin{figure*}
\centering
\includegraphics[width=6.5in]{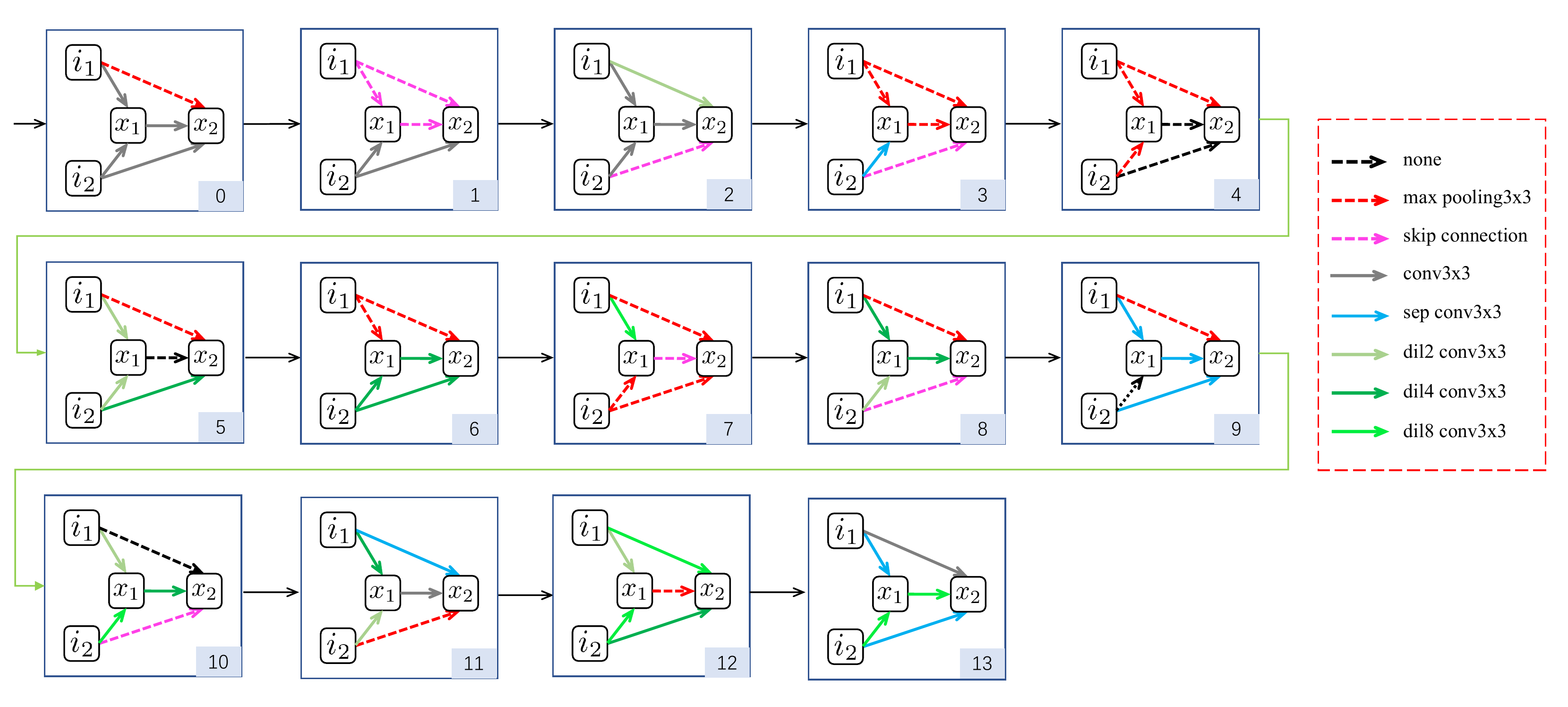}
\caption{The network searched by our GAS with GGM exhibits the benefit property (e.g. more dilation convolution operations in deep layers and more low computational operations for fast speed) for real-time semantic segmentation.}
\label{gas_network}
\end{figure*}

\begin{figure*}
\centering
\includegraphics[width=6.5in]{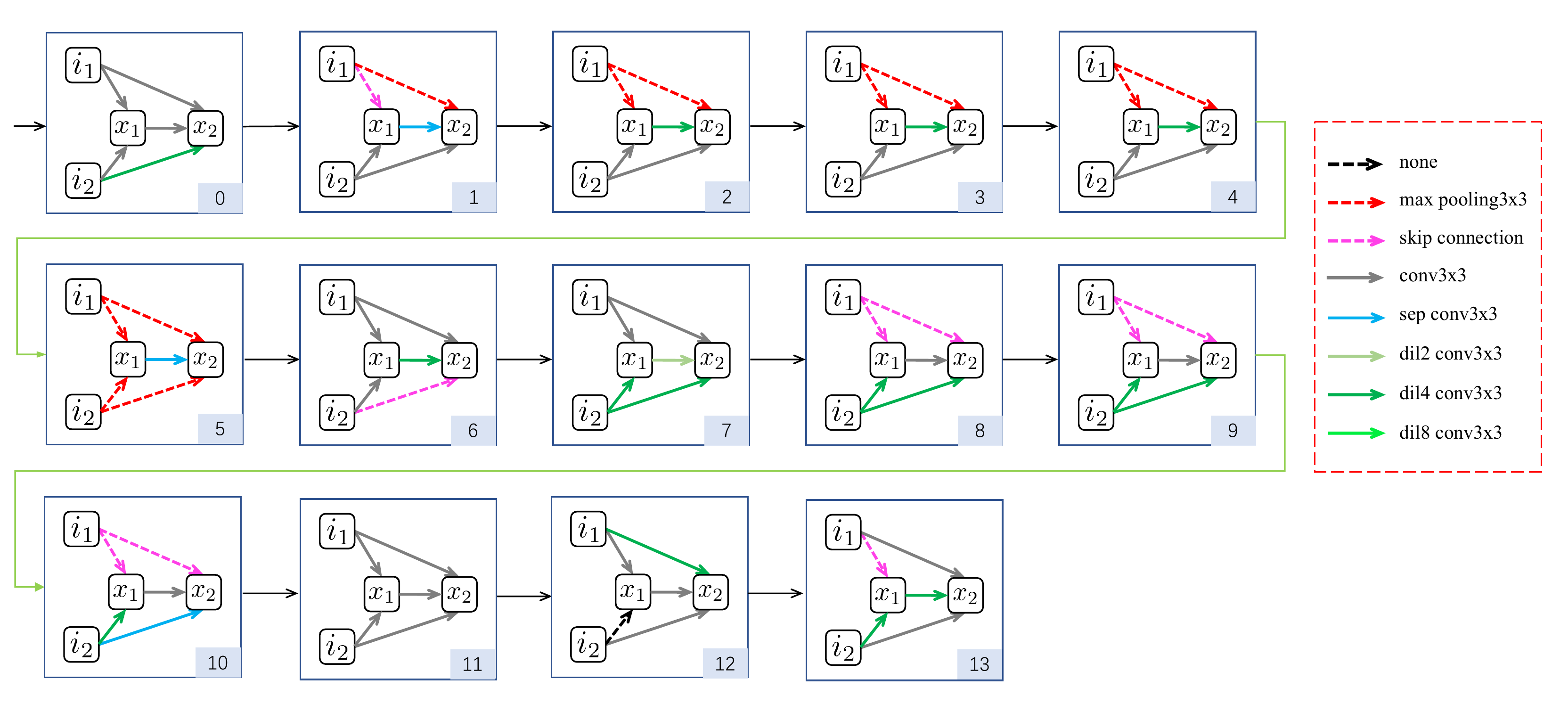}
\caption{The network searched by our GAS with fully connected layer.}
\label{fc_network}
\end{figure*}

\begin{figure*}
\centering
\includegraphics[width=6.5in]{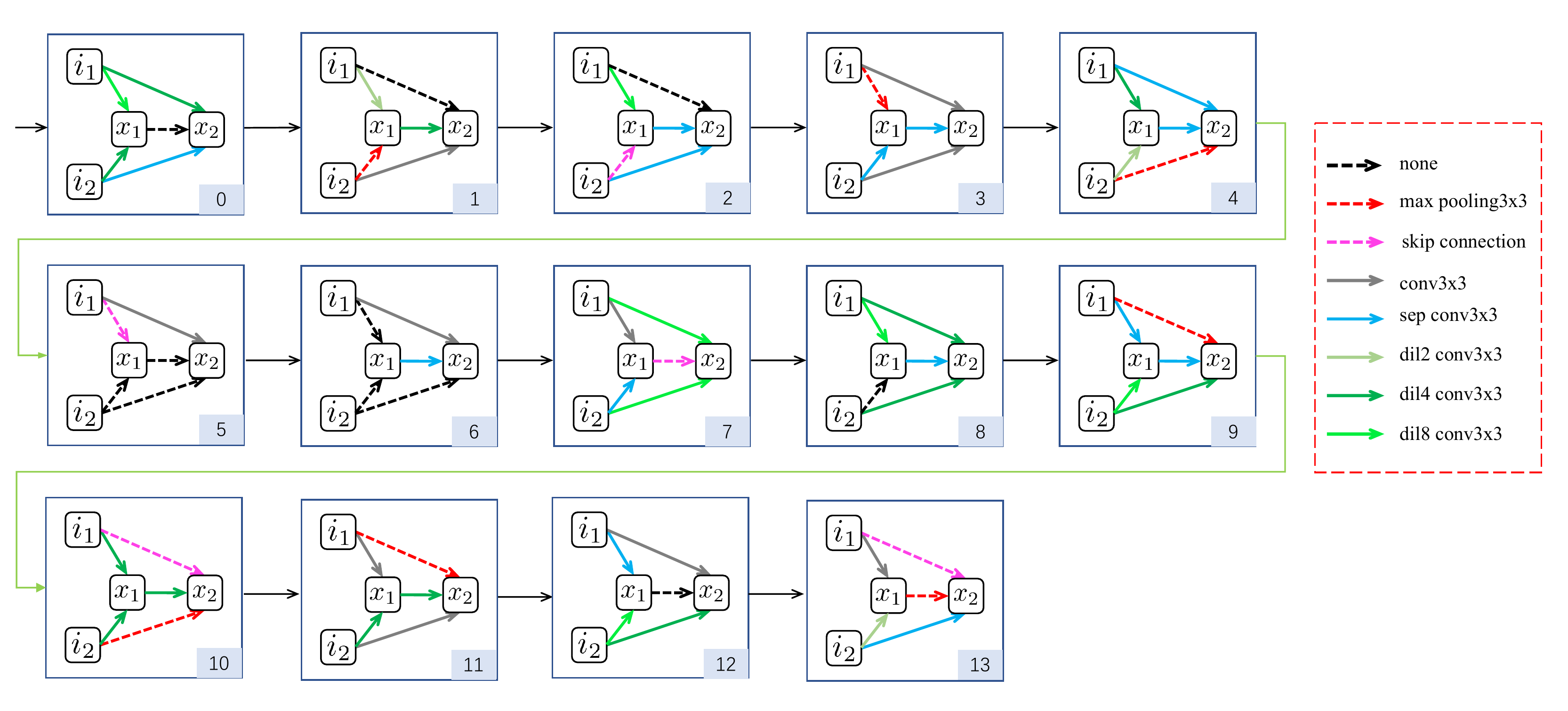}
\caption{The random searched network.}
\label{random_network}
\end{figure*}

As shown in Section 4.4.1 of the paper, the network searched by our GAS with GGM has smaller parameter size while achieving much higher performance. The visualization result can effectively help to analyze which component brings in the performance improvement. We thus visualize the networks searched by the three methods: 1) GAS with GGM; 2) GAS with fully connected layer; and 3) random search in Figure \ref{gas_network}, Figure \ref{fc_network} and Figure \ref{random_network}, respectively.

Compared to the other methods, the network searched by our GAS with GGM shows the following three advantages:

1) The cells in the low stage tend to choose light-weight operations (i.e. none, max pooling, skip connection) and the cells in the high stage enjoy the complex ones, which is the goal of pursuing high speed as described in the introduction of our paper. Specifically, under the same latency loss weight, the network searched by our GAS with GGM contains thirty light-weight operations (dashed-line arrow in the picture) with lower latency, while the other two methods use twenty-one and twenty-three light-weight operations, respectively. However, our GAS with GGM achieves higher performance, which exhibits the emergence of the concept of burden sharing in a group of cells when they know how much others are willing to contribute.

2) The deeper layers tend to utilize larger receptive field operations (e.g. conv with dilation = 4 or 8), which plays a key role to improve performance in semantic segmentation \cite{chen2018deeplab,DBLP:journals/corr/ChenPSA17}. Specifically, the network searched by our GAS with GGM uses 11 large receptive field operations (denoted by green arrow) in the last four cells and the other methods only use 4 or 8 operations, respectively.

3) The final structure has sufficient cell-level diversity as we expected. On the contrary, the network search by GAS with fully connected layer tends to use similar structures, for example, cell 7 is similar to cell 8 and 9, and cell 1 is similar to the cell 2, 3 and 4.

\section{Multi-Scale Module Exploration}

\begin{table}[h]
\begin{center}
\setlength{\tabcolsep}{1mm}{
\scalebox{0.9}{
\begin{tabular}{|l|c|c|}
\hline
Methods  & mIoU (\%) & FPS \\
\hline\hline
ASPP  & 72.4  & 108.4  \\
PPM & 72.5  & 114.1  \\
\hline
\end{tabular}}
}
\end{center}
\caption{The performance for different multi-scale modules on the Cityscapes validation set.}
\label{ppm}
\end{table}

When considering multi-scale features, we also try the PPM module in PSPNet \cite{zhao2017pyramid}, and our GAS achieves the similar performance with faster speed on the Cityscapes validation set in Table \ref{ppm}.

\end{document}